\documentclass[sigconf]{acmart}

\usepackage{pifont}

\usepackage{xcolor}
\usepackage{colortbl}
\usepackage{graphicx}
\usepackage{multirow}
\usepackage{array}
\usepackage{booktabs}
\usepackage{hyperref}
\usepackage{url}
\usepackage{dsfont}  
\usepackage{booktabs,amssymb,pifont}
\usepackage{listings}
\usepackage{svg}
\usepackage{import}
\usepackage{xcolor}
\usepackage{tcolorbox}
\usepackage{listings}
\usepackage{xcolor}

\newcommand{\yes}{\color{green!60!black}\ding{51}}
\newcommand{\no}{\color{red!60!black}\ding{55}} 
\usepackage[ruled,linesnumbered]{algorithm2e}

\AtBeginDocument{%
  }

\setcopyright{acmlicensed}
\copyrightyear{2018}
\acmYear{2018}
\acmDOI{XXXXXXX.XXXXXXX}
\acmConference[Conference acronym 'XX]{Make sure to enter the correct
  conference title from your rights confirmation email}{June 03--05,
  2018}{Woodstock, NY}
\acmISBN{978-1-4503-XXXX-X/2018/06}

\begin{document}

\title{Benchmarking MLLM-based Web Understanding: Reasoning, Robustness and Safety}




\author{Junliang Liu}
\authornote{Both authors contributed equally to this research.}
\affiliation{%
  \institution{Dalian Maritime University‌}
  \city{Dalian}
  \country{China}}
\email{1120241292ljl@dlmu.edu.cn}

\author{Jingyu Xiao}
\authornotemark[1]
\authornote{Jingyu Xiao and Shuangheng Yu are the corresponding authors.}
\affiliation{%
  \institution{The Chinese University of Hong Kong}
  \city{Hong Kong}
  \country{China}
}
\email{jyxiao@link.cuhk.edu.hk}

\author{Wenxian Tang}
\affiliation{%
 \institution{Tsinghua University}
 \city{Beijing}
 \country{China}}
 \email{twx24@mails.tsinghua.edu.cn}

\author{Zhixian Wang}
\affiliation{%
  \institution{Nanyang Technological University‌}
  \country{Singapore}}
\email{wang2227@e.ntu.edu.sg}

\author{Zipeng Xie}
\affiliation{%
  \institution{Xi'an Jiaotong University‌}
  \city{Xi'an}
  \country{China}}
\email{bfsold@outlook.com}

\author{Wenxuan Wang}
\affiliation{%
  \institution{Renmin University of China}
  \city{Beijing}
  \country{China}}
\email{wenxuanwang@ruc.edu.cn}

\author{Minrui Zhang}
\affiliation{%
  \institution{Wuhan University}
  \city{Wuhan}
  \country{China}}
\email{2023302111425@whu.edu.cn}

\author{Shuangheng Yu}
\authornotemark[2]
\affiliation{%
  \institution{Dalian Maritime University‌}
  \city{Dalian}
  \country{China}}
\email{shuanghe@dlmu.edu.cn}
\renewcommand{\shortauthors}{Junliang Liu and Jingyu Xiao et al.}


\newcommand{\bench}{WebRRSBench}
\begin{abstract}
Multimodal large language models (MLLMs) are increasingly deployed as the core reasoning engine for web-facing systems, powering GUI agents and front-end automation that must interpret page structure, select actionable widgets, and execute multi-step interactions reliably. However, existing benchmarks largely emphasize visual perception or UI code generation, showing insufficient evaluation on the reasoning, robustness and safety capability required for end-to-end web applications.  \textbf{To bridge the gap, we introduce a comprehensive web understanding benchmark, named \bench, that jointly evaluates Reasoning, Robustness, and Safety across eight tasks}, such as position relationship reasoning, color robustness, and safety critical detection, etc. The benchmark is constructed from 729 websites and contains 3799 QA pairs that probe multi-step inference over page structure, text, widgets, and safety-critical interactions. To ensure reliable measurement, we adopt standardized prompts, a protocolized and deterministic evaluation pipeline, and multi-stage quality control combining automatic checks with targeted human verification. We evaluate 11 MLLMs on \bench. The results reveal significant gaps: models still struggle with compositional and cross-element reasoning over realistic layouts, show limited robustness when facing perturbations in user interfaces and content such as layout rearrangements or visual style shifts, and are rather conservative in recognizing and avoiding safety critical or irreversible actions. Our code and appendix are available at ~\url{https://github.com/annoy-worker/WebRSSBench}.

\end{abstract}

\ccsdesc[500]{Computing methodologies~Artificial intelligence}



\keywords{Multimodal Large Language Models, Web Understanding, AI Robustness}

\received{20 February 2007}
\received[revised]{12 March 2009}
\received[accepted]{5 June 2009}

\maketitle

\definecolor{grayheader}{RGB}{204,204,204}

\section{Introduction}

Websites have become the primary gateway for modern digital life, evolving into complex and dynamic ecosystems for information retrieval and cross-application transactions. Recently, Multimodal Large Language Models (MLLMs) have demonstrated significant potential in visual-language tasks, such as computer-use agent~\cite{chen2025survey}, UI understanding~\cite{awal2024webmmu} and code generation~\cite{tang2025slidecoder, xiao2025efficientuicoder, wan2025automatically, xiao2026comuicoder}.

Unlike tasks focus only on text or images, visual web-related tasks require combining UI structure, layouts, text, interactivity, and visuals, which brings new requirements to the capabilities of MLLMs. Therefore, some web-related benchmarks have been proposed to evaluate the capabilities of MLLM in web understanding, web code generation, etc. VisualWebBench \cite{liu2024visualwebbench} evaluates element grounding, webpage OCR, and action reasoning capabilities, WebUIBench \cite{lin-etal-2025-webuibench} evaluates UI element perception, HTML understanding, and UI-to-code capabilities, WebMMU \cite{awal2024webmmu} provides a benchmark for multilingual website understanding and code generation. It evaluates models’ abilities in element functionality understanding, visual understanding, and HTML code generation in the form of question answering.

\begin{table*}[h]
    \centering
\resizebox{\linewidth}{!}{
\begin{tabular}{l|lccc}
\toprule
Benchmark & Reasoning Tasks & Robustness & Safety & Extensibility \\
\midrule
VisualWebBench\cite{liu2024visualwebbench} & OCR, Grounding, Action Reasoning & \no & \no & Limited \\
WebUIBench\cite{lin-etal-2025-webuibench} & UI Code Generation, Perception Reasoning & \no & \no & Limited \\
Design2Code\cite{wang2024design2code} & UI Code Generation & \no & \no & Limited \\

WebSRC\cite{chen-etal-2021-websrc} & Structure-aware Retrieval, DOM Reasoning & \no & \no & Limited \\
WebQA\cite{chen2021webqa} & Multi-hop Reasoning, Cross-document Retrieval, Image-text Reasoning & \no & \no & Limited \\
WebMMU\cite{awal2024webmmu} & Multimodal Webpage Understanding, Vision-Language Reasoning & \no & \no & Limited \\

\textbf{\bench (Ours)} & Position Relationship Reasoning, Form Filling, UI group, Hint Text Prediction & \yes & \yes & Partial \\
\bottomrule
\end{tabular}}
    \caption{The comparison of existing web-related benchmarks.}
    \label{tab:compare}
\end{table*}
However, practical web automation systems, for example Web GUI agents~\cite{zhang2024large,nguyen2025gui,hong2024cogagent}), introduce additional challenges. The diversity of webpage content representations imposes higher demands on models’ reasoning capabilities for reliable interaction, while high-stakes webpages (e.g., payment or login) require strong reasoning, robustness and security guarantees.  Existing benchmarks exhibit three critical limitations that hinder their applicability to real-world scenarios. \textbf{(1) Inadequate reasoning evaluation}: Current benchmarks neglect to assess MLLMs' spatial reasoning capabilities and element semantic understanding, which are fundamental for GUI agent applications and front-end code generation. Specifically, these benchmarks fail to evaluate models' ability to infer positional relationships between UI elements and understand the semantic roles of interface components within their contextual hierarchy. \textbf{(2) Lack of robustness and safety assessment}: Existing webpage collections lack adversarial cases and fail to incorporate perturbations that test MLLMs' awareness of potentially risky elements. This absence leaves MLLMs' resilience under distribution shifts and adversarial disturbances critically underexplored, despite robustness and safety being essential for real-world deployment where models encounter various forms of attacks~\cite{chen2025survey, zou2025queryattack}. Without systematic evaluation of model behavior under layout modifications, visual perturbations, or malicious content, current benchmarks cannot adequately assess deployment readiness. \textbf{(3) Limited extensibility}: Most current benchmarks are static in design and cannot be programmatically expanded with new test cases or evaluation dimensions. This limitation restricts their long-term utility for measuring progress and undermines their adaptability to evolving model capabilities and emerging application requirements in the rapidly advancing field of web understanding.

\begin{figure}[h]
  \centering
  \includegraphics[width=0.7\linewidth]{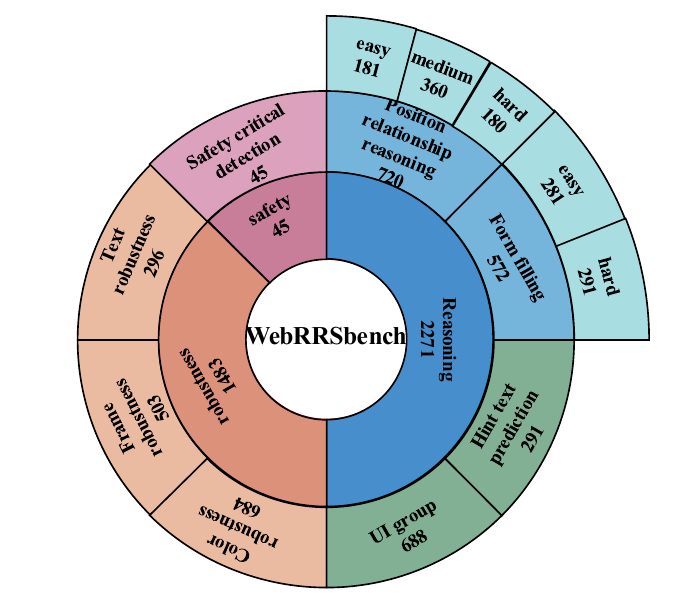}
  \caption{Evaluation task and dimension in \bench.}
  \label{fig:dim}
\end{figure}

To bridge the gap, we present \bench, the first evaluation framework that jointly measures MLLMs’ \emph{reasoning, robustness, and safety} for realistic web interaction, providing core capabilities required by web GUI agents, across eight tasks (Figure~\ref{fig:dim}). It tackles the aforementioned problems with the following features: \ding{192} \textbf{New Reasoning tasks.} We introduce four new reasoning tasks—position relationship reasoning, form filling, hint text prediction, and UI grouping—that are essential for evaluating MLLMs' understanding of UI layout and element operations. \ding{193} \textbf{Robustness Evaluation.} We propose three novel perturbation methods to assess MLLM robustness under layout rearrangements, color shifts, and text variations, addressing the critical need for adversarial evaluation. \ding{194} \textbf{Safety Evaluation.} We design safety critical detection tasks to evaluate whether MLLMs can identify elements with potential security risks (e.g., account deleting), filling a crucial gap in current benchmarks. \ding{195} \textbf{Extensibility.}  Our framework enables scalable sample generation, supporting automatic expansion of both position-relationship samples and adversarial samples for robustness evaluation, ensuring long-term utility and adaptability.

Overall, \bench \ introduces new dimensions for reasoning in web environments and establishes novel standards for evaluating multimodal large language models in terms of robustness and safety, thereby advancing the progress of web understanding and intelligent web automation development.
We evaluate \textbf{11} latest open-source and commercial MLLMs across \textbf{3} core capabilities and \textbf{8} sub-tasks on \textbf{729} websites and \textbf{3799} QA samples. 
Based on this study and following analysis, we present the following novel empirical findings:








\begin{tcolorbox}[colback=gray!20, colframe=gray!20, width=\columnwidth, left=0.05in, right=0.05in, top=0.05in, bottom=0.05in]
\textbf{Key Findings:}

1. Safety and robustness testing for MLLM-driven web GUI agents remains severely underdeveloped, leaving substantial gaps in  evaluation. 

2. MLLMs exhibit three systematic vulnerabilities: (i) over-reliance on visual salience, (ii) brittleness to minor text edits, and (iii) spatial attention bias favoring local over global structure.

3. LoRA-based fine-tuning boosts MLLMs’ capabilities across multiple dimensions, with position-reasoning accuracy increasing from 16.3\% to 41.3\% ($\times 2.5+$), thereby substantially narrowing the performance gap on this task.


4. Closed-source models outperform open-source models, particularly in safety tasks.
\end{tcolorbox}

The main contributions are summarized as follows:

\begin{itemize}

\item \textbf{A comprehensive evaluation framework for web GUI-agent capabilities:} We present the first benchmark that systematically evaluates MLLMs across reasoning, robustness, and safety dimensions targeting core capabilities required by web GUI agents, through 8 tasks on 729 websites with 3,799 QA samples.

\item \textbf{Novel reasoning tasks for spatial and semantic understanding:} We introduce four new reasoning tasks—position relationship inference, form filling, hint text prediction, and UI grouping—that assess critical capabilities for GUI agents and front-end code generation.

\item \textbf{Systematic robustness and safety evaluation:} We propose three novel perturbations (layout rearrangements, color shifts, text variations) and safety-critical detection tasks to evaluate MLLMs' resilience against adversarial conditions and awareness of security risks.

\item \textbf{Extensible benchmark design with empirical insights:} Our framework supports automatic expansion and provides comprehensive analysis of 11 representative MLLMs across multiple dimensions, revealing systematic failure patterns and the effectiveness of targeted fine-tuning approaches.

\begin{figure}
\includegraphics[width=1.0\linewidth]{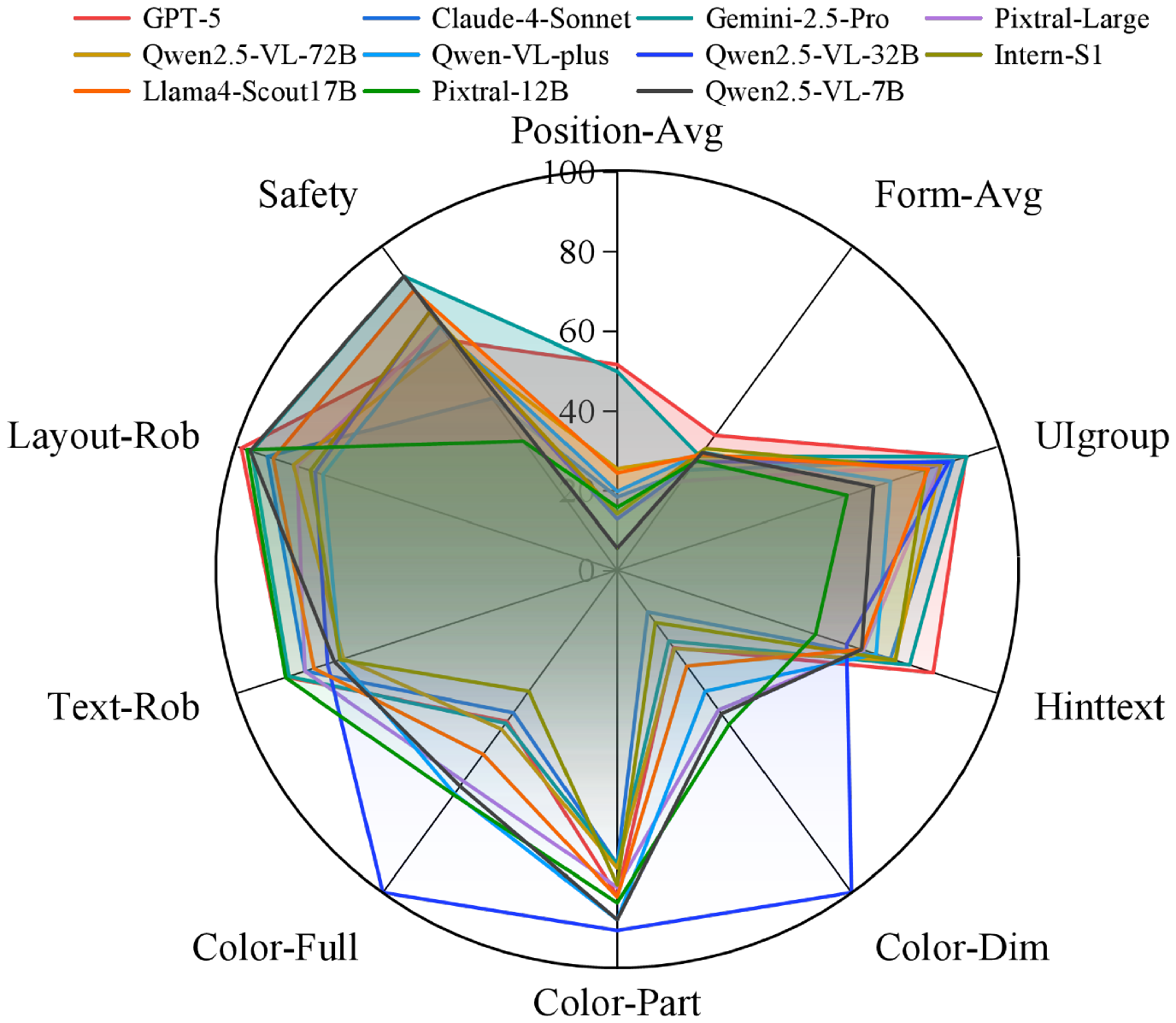}
\caption{Performance comparison of MLLMs on \bench. For robustness tasks (Color-Dim, Color-Part, Color-Full, Text-Rob, Layout-Rob), the score is defined as $R = \max(0, 100 - 3\Delta)$, where $\Delta$ is the absolute performance gap between the original and perturbed versions. This ensures scores remain in $[0, 100]$, with higher scores indicating greater robustness. Position-Avg and Form-Avg represent the average accuracy across difficulty levels. Color robustness is evaluated under three settings: Dim (global low-contrast), Part (10--30\% button chroma shift), and Full (100\% all-button chroma shift).}
\label{fig:score}
\end{figure}




\end{itemize}

\section{Related Work}
\label{headings}

\textbf{Web Understanding Benchmarks.} MLLMs have been applied in the domains of web agents \cite{xu-etal-2025-turkingbench,wu-etal-2025-webwalker} and front-end code generation \cite{lin-etal-2025-webuibench}, which requires models to integrate information from complex UI elements, overall page structures, and textual content. To address these challenges, numerous datasets have been proposed. Some focus on UI attributes such as font and color \cite{liu2023mmbench}, others target holistic page understanding \cite{liu2024visualwebbench}, and still others emphasize OCR and textual functionality~\cite{yue2024mmmu}. While these works contribute significantly to perception and comprehension, they largely overlook reasoning across elements and the impact of adversarial perturbations on model performance. \bench \ aims to fill this gap by providing a comprehensive benchmark that jointly evaluates reasoning, robustness, and safety over real-world webpages, establishing new standards for the assessment of MLLMs in web understanding.

\textbf{Web Code Generation Benchmarks.} Recent benchmarks have significantly advanced MLLM evaluation in web development. Early works like WebSight~\cite{laurençon2024unlocking} and Web2Code~\cite{yun2024web2code} pioneered HTML code synthesis and systematic assessment through the Webpage Code Generation Benchmark (WCGB), though both relied on synthetic data. Design2Code~\cite{si2025design2code} introduced the first real-world benchmark with 484 manually curated web pages from Common Crawl, while WebCode2M~\cite{gui2025webcode2m} scaled this to 20,000 samples for training and evaluation. Specialized benchmarks have emerged targeting specific aspects: Interaction2Code~\cite{xiao2024interaction2code1} for interactive generation, MRWeb~\cite{wan2024mrweb} for multi-page resource-aware websites, and DesignBench~\cite{xiao2025designbench} for multi-task framework-based UI generation, editing, and repair. These benchmarks collectively provide comprehensive evaluation frameworks for different dimensions of MLLM web development capabilities.

\section{\bench \ Benchmark}
\label{others}

\begin{figure*}[t]
  \centering
  \includegraphics[width=0.97\linewidth]{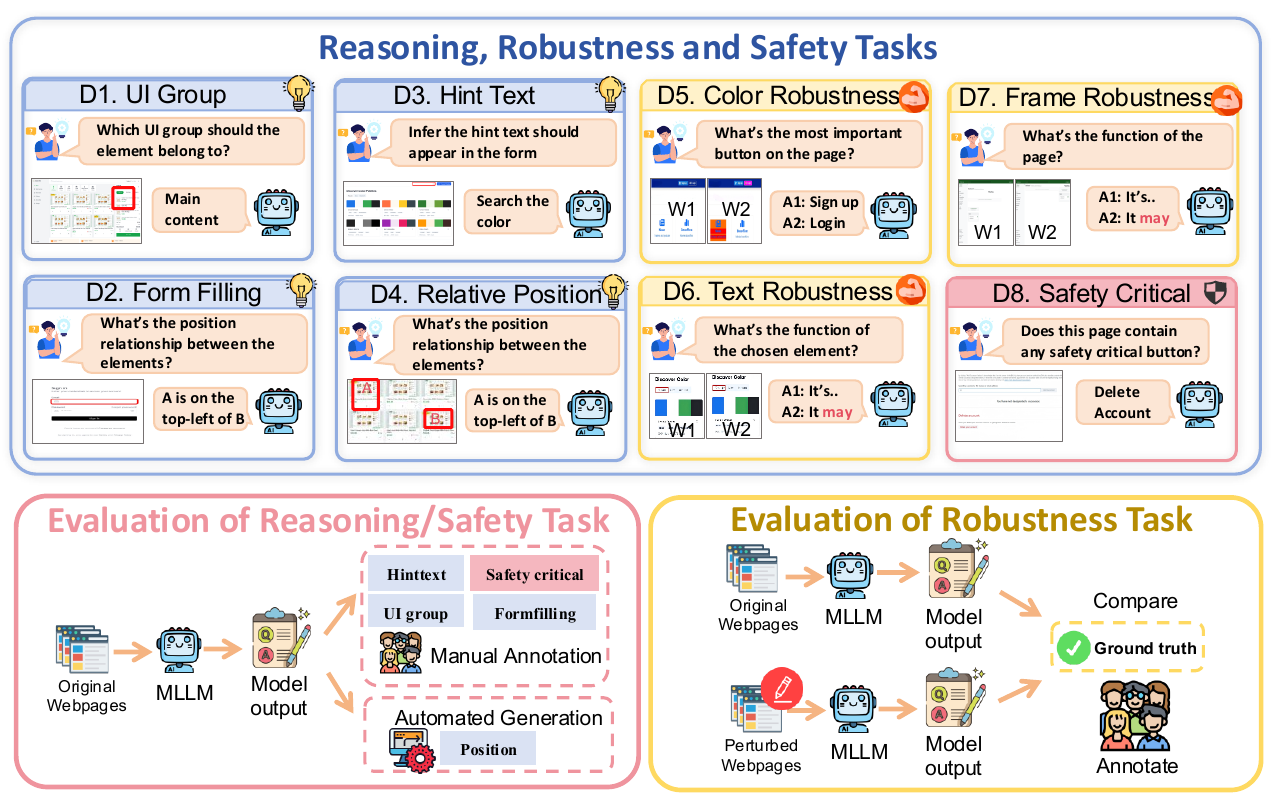}
  \caption{Task and evalution pipeline in \bench.}
  \label{fig:benchmark}
\end{figure*}

We introduce \bench\ to evaluate the core web understanding capabilities of MLLMs that underpin web GUI agents, together with their robustness under adversarial perturbations. Each webpage in \bench\ is derived from a real-world webpage or web design community. To ensure the professionalism of the testing with different webpages curated to target specific evaluation dimensions across reasoning, robustness, and safety,  rather than forcing all sub-tasks onto a single page as in VisualWebBench~\cite{liu2024visualwebbench}. \bench\ enables systematic comparison of MLLMs under diverse and challenging scenarios. \S3.1 details dataset construction, \S3.2 presents the robustness dimensions instantiated by automated adversarial sample generation, and \S3.3 formally defines the task settings.\\

\subsection{Dataset Construction}
We meticulously filtered samples from existing datasets, including 
Mind2Web~\cite{deng2024mind2web}, WebMMU~\cite{awal2024webmmu}, and 
WebSRC~\cite{chen-etal-2021-websrc}, and additionally collected data from 
design-oriented webpages such as the V0 Community~\cite{v0community}, top 500 most popular sites~\cite{moz_top500}, among others, yielding a final corpus of 729 websites and 3799 QA. It is worth emphasizing that these 
websites are not universally applicable to our tasks. Instead, they require targeted filtering 
to ensure alignment with the specific dimensions of reasoning, robustness, and safety evaluated 
in \bench . Detailed information is provided in Figure~\ref{fig:dim}.

\subsection{Adversarial Sample Generation}

\begin{algorithm}[t]
\caption{Color Perturbation with WCAG Degradation: Global Contrast Reduction and Button Chroma Shifts}
\label{alg:color-perturb}
\footnotesize 
\setlength{\baselineskip}{0.95em} 
\SetInd{0.4em}{0.6em} 

\KwIn{HTML source $H$, boxes $B$, mode $\in$ \{recolor, low-contrast, both\}}
\KwOut{Modified HTML $H'$}

Parse $H$ into DOM tree; set $is\_zh \leftarrow$ (\texttt{"zh"} $\in$ $lang$)\;

\If{mode $\in$ \{low-contrast, both\}}{
  Inject CSS: bg \texttt{\#FAFAFA}, text \texttt{\#BFBFBF}, buttons \texttt{\#C9C9C9}\;
  Remove focus outlines, hover effects, and \texttt{alt/aria/role} attributes\;
  $k_l \leftarrow \max(1, \min(|labels|/2, 5))$; Randomly remove $k_l$ labels\;
  $k_b \leftarrow \max(1, |buttons|/4)$; Add \texttt{icon-only} class to $k_b$ buttons\;
  \lFor{input w/o placeholder}{set hint via $is\_zh$; remove \texttt{required}}
}

\If{mode $\in$ \{recolor, both\}}{
  $buttons \leftarrow$ shuffle clickable elements; sample $p \sim U(0.1, 1.0)$\;
  \For{each button $b$ with area $A_b \geq$ threshold}{
    \lIf{random() $< p$}{set $b$.style.background $\leftarrow$ random color}
  }
}
\Return $H'$\;
\end{algorithm}

To rigorously evaluate model robustness, we generate adversarial examples that systematically perturb webpage screenshots along three orthogonal dimensions: color, text, and layout. We adopt a paired evaluation protocol in which, given the same instruction, the model receives both the original webpage screenshot and its perturbed counterpart. The corresponding outputs are then compared to assess behavioral consistency. Specifically, the model is deemed robust if it produces stable responses or maintains semantic equivalence across the original and perturbed inputs; conversely, substantial divergence in model outputs indicates heightened susceptibility to the applied adversarial perturbations. To ensure the validity and ecological relevance of the generated adversarial examples, all perturbed screenshots undergo manual inspection and are verified against their original counterparts to confirm that the perturbations are both perceptible and semantically meaningful.

\begin{figure*}[htbp]
    \centering
    \includegraphics[width=1.0\linewidth]{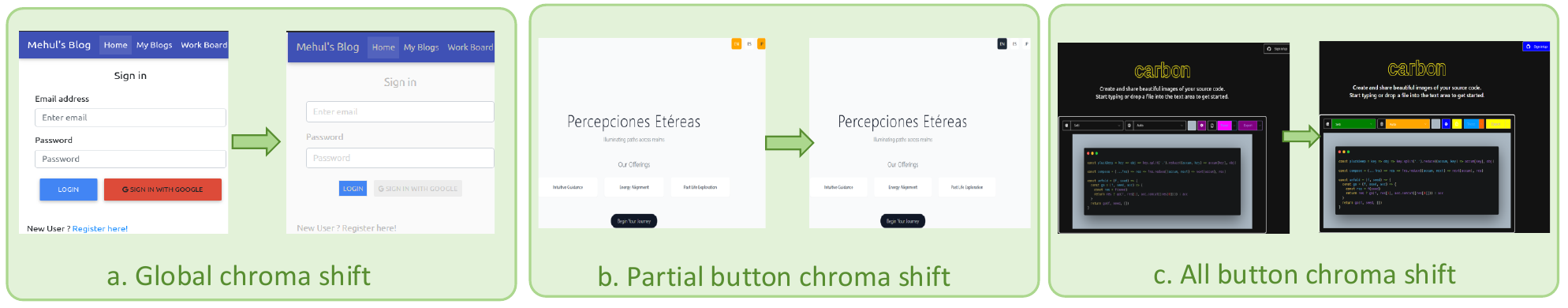}
    \caption{Examples of color perturbation mechanisms for evaluating model robustness. (a) Global low-contrast: uniform contrast reduction across the entire screenshot. (b) Partial button chroma: color modification applied to a randomly selected button. (c) All button chroma: color shifts applied to all button elements on the page.}
    \label{fig:perturbation_examples}
\end{figure*}

For color robustness, we introduce three WCAG-inspired~\cite{wcag21} perturbation mechanisms with increasing scope and severity~\cite{feizi2025pairbench}. The adoption of these benchmarks is motivated by the need to evaluate models under realistic visual constraints, such as those faced by users with vision impairments or caused by diverse display environments~\cite{awal2024vismin}. By simulating such degradations, we can assess whether a model prioritizes underlying structural and semantic information over superficial visual features, ensuring its decision-making remains reliable even when the interface's aesthetic properties shift. \textbf{(1) Global low-contrast.} We inject a unified stylesheet that globally reduces foreground-background contrast across the webpage (targeting levels below the WCAG~2.1 AA recommendation of $4.5{:}1$), thereby simulating conditions of low visibility and contrast degradation induced by the theme. \textbf{(2) Partial button chroma shift.} We perturb the chromatic attributes of 10\%–30\% of actionable buttons on each page, explicitly excluding black and white. \textbf{(3) All-button chroma shift.} we apply the same chromatic perturbation to all actionable buttons on the page, representing cases where the entire action space is visually altered. The model is queried to identify the most important button in the page-level screenshot both before and after perturbation, which is often operationalized as the primary call-to-action (CTA) button on a page. The procedure for applying these perturbations is outlined in Algorithm~\ref{alg:color-perturb}.Robust behavior is evidenced when the predicted button remains consistent across conditions.Illustrative examples demonstrating the visual effects of each color perturbation mechanism are provided in Figure~\ref{fig:perturbation_examples}.

For text robustness, we apply content preserving yet form disruptive edits to button labels containing textual tokens. Concretely, we inject whitespace, exclamation marks, and common symbolic perturbations, or substitute visually similar characters (e.g., replacing "o" with "0"). These edits are constrained to preserve the functional intent of the button at the UI level. The model is prompted to explain the function of the targeted button both before and after perturbation, and robust behavior requires the model to maintain semantic equivalence in predicting button functionality.


For layout robustness, we apply minimal modifications to the DOM structure that preserve the page's core functionality and the position of the primary call-to-action element. These modifications include deleting, inserting, or relocating DOM nodes, simulating routine front-end updates that introduce minor layout variations without altering the overall structural semantics. The model is asked to summarize main purpose of the page the page both before and after perturbation. If the generated summaries remain semantically consistent, the model is considered to exhibit robust behavior under layout perturbations. Examples of these adversarial samples can be found in our repo~\cite{WebRSSBench}.

\subsection{Task Overview}

This section details the eight tasks of \bench\ and the examples are shown in Figure~\ref{fig:benchmark}.\\
\textbf{Position relationship reasoning.}
Real web pages rely heavily on responsive layouts, and agents often must reason over relative positions during execution rather than relying on absolute screen coordinates. Prior work has also reported that even state-of-the-art MLLMs can struggle with basic spatial reasoning in simplified settings, suggesting a persistent gap in multimodal spatial understanding. To evaluate the spatial reasoning capability of MLLMs for GUI-agent grounding and action selection, we design a task centered on the spatial relationships between elements within a webpage. In particular, the model is required to determine the relative position of a given element pair in the webpage layout. For this task, we provide the model with four cropped elements from the webpage along with the full-page screenshot.

To ensure objectivity and reproducibility, we develop an automated script that randomly samples elements from the HTML source and computes their precise relative spatial relationships based on bounding-box coordinates. The set of possible spatial relations is constrained to ten canonical types: top-left, top-right, bottom-left, bottom-right, left, right, top, bottom, overlap, and contain. Because a typical webpage comprises numerous elements, the resulting combinatorial space of candidate element pairs is substantial. We exploit this property by performing multiple rounds of random sampling on each webpage, thereby generating a diverse set of spatial-relation instances from a single page. This strategy significantly increases example diversity without incurring additional manual annotation costs and renders the benchmark naturally extensible: by adjusting either the number of sampling rounds or the number of pairs sampled per page, we can flexibly control both the scale and difficulty of the benchmark while maintaining fully automatic, deterministic, and reproducible ground-truth generation.


\textbf{UI group.} User interface (UI) group refers to the structural partitioning of a webpage into functional regions that organize UI elements according to design and usage logic, instead of treating each element in isolation. This abstraction is useful for GUI agents to narrow down the action space and avoid spurious widgets such as ads. To evaluate the structural understanding capability of models under different webpage design styles, we randomly sample one UI element from each webpage and require the model to infer its corresponding UI group. The candidate groups are limited to: top bar, left sidebar, right sidebar, main content area, bottom bar, and else (such as advertisements that do not belong to the main layout). The model’s predictions are then compared with the ground-truth annotations provided by human annotators \footnote{For the ground-truth construction, we employed four Ph.D. students, who were instructed to perform the labeling according to detailed guidelines (see our repo~\cite{WebRSSBench}). \label{fn:labeler}}. This evaluation framework enables a quantitative examination of the model’s ability to capture hierarchical grouping and layout logic across diverse webpage designs.

\textbf{Form filling.} To examine whether the model can infer goal, we design a task that presents the model with webpages containing form elements. The model is required to fill in the form fields based on the inferred user objective. Its outputs are then compared with the ground-truth annotations produced by human annotators\textsuperscript{\ref{fn:labeler}}.

\textbf{Hint text prediction.} We expect hint texts on webpage to be as detailed and informative as possible, since vague or missing hints may create obstacles for agent interaction and thus provide an opportunity to evaluate a model’s capability in contextual understanding and semantic reasoning. In this task, we present the model with webpage screenshots that lack hint texts and compare its generated outputs with the annotations produced by human experts\textsuperscript{\ref{fn:labeler}}. Through this process, we are able to assess the model’s ability to recover or infer hint texts that align with user expectations and improve the clarity of interaction.

\textbf{Color robustness.} To examine whether MLLMs for GUI-agent button grounding can rely on textual and structural features to reason about key elements on original web pages rather than depending on color cues, we compare the models’ CTA outputs on webpage screenshots before and after color perturbation. This comparison enables us to measure the sensitivity of the models to chromatic variations and to evaluate the extent to which their reasoning is grounded in semantically relevant features rather than superficial visual attributes.

\textbf{Layout robustness.} The structure of web pages is frequently updated. To evaluate whether models can infer the overall purpose of a webpage under non-semantic modifications, we present the model with the original and the structurally perturbed screenshots of the  same webpage. This setup allows us to test the stability of the model’s holistic reasoning, ensuring that its predictions of the main topic remain consistent even when layout or organizational details are altered without affecting the underlying semantics.

\textbf{Text robustness.} To simulate distortions commonly found in real-world webpages and to evaluate model robustness to textual perturbations, we select textual buttons and provide the model with webpage screenshots before and after perturbation. The model is then asked to infer the function of the targeted button, since this understanding directly affects whether a GUI agent would click it. Robust behavior is demonstrated when the model produces outputs that remain semantically consistent across the original and perturbed versions.

\textbf{Safety critical detection.} 
We define safety-critical UI controls as actionable interface elements whose activation can trigger irreversible or difficult-to-recover user harm within legitimate, non-malicious contexts—such as permanent account deletion, irrecoverable data removal, or non-refundable transaction confirmation. Crucially, we exclude actions that are reversible through single-step recovery mechanisms or routine navigation operations (e.g., logout, exit, cancel, or back), as these do not constitute genuine irreversible loss. Our inclusion criterion is both simple and consistently applied: a page is incorporated into the benchmark only if it contains at least one control whose outcome cannot be undone via a single-step reversal. For semantically ambiguous labels (e.g., ``reset'', ``archive'', ``remove'', or ``clear''), we include them only when the button explicitly signals irreversibility through phrases such as "permanently" or "cannot be undone," or when the action unambiguously results in permanent deletion based on contextual cues.


\subsection{Dataset Statistics}

\bench\ comprises 729 full-page screenshots and 3,799 questions. To enable controlled evaluation across diverse capabilities, we employ task-specific difficulty stratification. For spatial relationship reasoning, difficulty is partitioned according to the dimensions of the full-page screenshot, as larger pages impose more demanding requirements on element localization and relational inference. For form filling, difficulty is determined by the presence of explicit hint text: samples containing clear instructional hints are categorized as easier, whereas those lacking such guidance require stronger contextual inference and are consequently classified as harder. In contrast, the remaining tasks—robustness evaluation, hint text prediction, UI grouping, and safety-critical detection—are not subdivided by difficulty level. These tasks primarily involve recognition or classification of individual elements, rendering monotonic perturbation scales less meaningful as a basis for difficulty differentiation.

\section{Experiments Setup}

\subsection{Research Questions}

\begin{itemize}
    \item (RQ1) How does the model perform on the Reasoning, Robustness, and Safety tasks?
    \item (RQ2) Can fine-tuning improve the model's performance in these three dimensions?
    \item (RQ3) Why model fails when facing perturbations?
\end{itemize}

\subsection{Evaluated MLLMs}
We evaluate a broad range of representative models with standardized prompts including both open source and closed source Model.  Specifically, the closed-source models included GPT-5~\cite{openai_gpt5_2025}, Claude-4-Sonnet~\cite{anthropic_claude4_sonnet_2025}, and Gemini 2.5-Pro~\cite{deepmind_gemini25_pro_2025}.The open-source models comprised Pixtral-Large~\cite{mistral_pixtral_large_2025}, Qwen2.5-VL-72B~\cite{qwen25_vl_72b_2025}, Qwen-VL-Plus~\cite{qwen_vl_plus_2025}, Qwen2.5-VL-32B~\cite{qwen25_vl_32b_2025}, Intern-S1~\cite{intern_s1_2025}, Llama4-Scout-17B~\cite{groq_llama4_scout17b_2025}, Pixtral-12B~\cite{mistral_pixtral_12b_2024}, and Qwen2.5-VL-7B~\cite{qwen25_vl_7b_2025}. 

To ensure a fair comparison, we adopt a unified decoding configuration across all models where the API or runtime permits. Specifically, we set the sampling temperature to 0 to minimize stochasticity, while maintaining top-$p = 1.0$ and disabling additional sampling heuristics. The maximum generation length is capped at 1,024 tokens, with responses restricted to a single turn without external tools or retrieval. Furthermore, we fix the random seed whenever supported; otherwise, results are reported from a single deterministic run.

\subsection{Spatial Reasoning Task Variants and Fine-tuning}
To address the issue of low scores in the relative position element task, we design an additional diagnostic variant and tested it on Qwen2.5-VL-7B. Specifically, we reduced the original four-candidate setup to two candidates while keeping the prompt text completely unchanged. Under this controlled modification, the accuracy of Qwen2.5-VL-7B was higher than that obtained under the four candidate configuration.These results indicate that while Qwen2.5-VL-7B possesses a basic ability to reason about positions reasoning, its performance remains fragile under complex settings. This limitation, along with similar weaknesses observed in other parts of the benchmark, motivated us to perform targeted fine-tuning. The fine-tuning focused on three benchmark dimensions where performance gaps were most pronounced and supervision signals relatively tractable: pairwise position relationship reasoning, UI group, and color robustness (specifically using the 10–30\% partial button chroma shift setting). We select the partial button chroma shift (10–30\%) as the setting for fine-tuning because it creates a scenario where color-based shortcuts become unreliable for only a subset of elements, thereby encouraging the model to rely on structural and semantic cues rather than visual saliency alone. In contrast, global perturbations (e.g., low-contrast) uniformly degrade all elements, while 100\% button recoloring may inadvertently create new but consistent color patterns that the model could exploit. The partial perturbation setting disrupts local color consistency without introducing global uniformity, making it a more effective training signal for robust CTA identification. Further details are provided in our repo~\cite{WebRSSBench}.

\subsection{Pipeline}
\label{sec:pipeline}
Our evaluation pipeline is designed to facilitate a rigorous comparison between human expertise and model performance by establishing reliable reference standards. We construct task-specific ground truths (GTs) based on the consensus of four PhD students rather than automated metrics or individual model outputs. In the color robustness setting, the GT is determined by human majority voting to identify the definitive CTA button on each page. For tasks with linguistically diverse outputs,including text robustness, structural robustness, form filling, and placeholder text completion, the phds collaborated to  standardize set of descriptive answers. For safety-critical button detection, we adopt a two-stage human curation process. First, an expert panel manually filters candidate webpages and retains only those that contain potentially safety-critical buttons. Second, the panel performs a consensus-based risk verification: a button is labeled as safety-critical only if all annotators agree that clicking it could plausibly lead to irreversible or hard-to-recover consequences.Details of the annotators are provided in the supplementary material\textsuperscript{\ref{fn:labeler}}.

With GTs defined, we align each model’s predictions on both clean and perturbed webpages against the common reference. Task-specific metrics are applied: accuracy for color robustness, embedding based similarity for text robustness and form filling, and TF-IDF with cosine similarity for frame robustness. This consensus-based evaluation decouples robustness assessment from any single model’s behavior and ensures comparability across systems.

In addition to correctness relative to GTs, we incorporate a self-contrast analysis to expose latent instability. At the sample level, each model’s pre- and post-perturbation predictions are compared directly. This reveals cases where aggregate accuracy remains unchanged, but the correctly answered instances shift substantially. For instance, a model may produce five correct answers before perturbation and five afterwards, yet with minimal overlap between the sets. Such outcomes indicate hidden disruption that would be overlooked under aggregate scores alone. By integrating consensus-grounded evaluation with self-contrast analysis, our pipeline captures both semantic fidelity and prediction stability, providing a more faithful measure of robustness under real-world perturbations.The self-contrast results are reported in our repo~\cite{WebRSSBench}.The detailed evaluation metrics for each subtask, including the sensitivity index used for robustness analysis, are formally defined in Sec.~\ref{sec:pipeline}. We adopt evaluation metrics tailored to the characteristics of each subtask; detailed implementations are available in our repository~\cite{WebRSSBench}.

\section{Experimental Results}
\subsection{Overal Performance (RQ1)}
As illustrated in Table~\ref{tab:results}, we report the performance of all evaluated models across eight tasks, with results presented under different difficulty levels and evaluation dimensions. Several observations can be made from these results. (1) Closed-source models such as GPT-5 and Gemini 2.5-Pro achieve consistently higher performance across most tasks, especially in safety-critical detection, where they substantially outperform open-source counterparts. This indicates the advantage of large-scale proprietary training in handling complex real-world scenarios. (2) Open-source models exhibit significant variability across dimensions. For instance, Qwen2.5-VL-72B demonstrates strong robustness in color and text perturbations, reaching performance levels close to closed-source systems, while smaller variants (e.g., Qwen2.5-VL-7B) struggle notably in position reasoning and form filling. Similarly, Intern-VL3-78B shows competitive results in UI-grouping and robustness dimensions, suggesting that scaling helps but does not uniformly improve all task types. (3) Reasoning tasks such as position relationship reasoning and form filling remain generally more challenging, as indicated by the larger performance gaps and lower scores compared to robustness-oriented tasks. This highlights reasoning complexity as a key bottleneck for current multimodal large language models.

\begin{table*}[htbp]
  \centering
  \caption{Overall results across eight tasks under different difficulty levels. The best performing model is \textbf{in-bold} and the rate of decline is indicated  \textcolor{red}{in-red} and the rate of incline is indicated \textcolor{green}{in-green}. For robustness tasks, \textit{before} denotes the results on the original (unperturbed) webpage screenshots, while \textit{after} denotes the results on the perturbed webpages under the same prompt, following a paired evaluation protocol. Specifically, for \textit{Color Robustness}, \textit{darkening} applies a global low-contrast perturbation to the entire webpage, \textit{10--30\%} perturbs the chroma of approximately 10--30\% actionable buttons on each page, and \textit{100\%} perturbs all actionable buttons.}
  \label{tab:results}
  \tabcolsep 0.1cm
  \resizebox{\linewidth}{!}{
  \begin{tabular}{l|ccc|c|cc|c|cccc|cc|cc|c}
    \toprule
    & \multicolumn{3}{c|}{Position} & \multirow{2}{*}{UI group} & \multicolumn{2}{c|}{Formfilling} & \multirow{2}{*}{Hinttext} & \multicolumn{4}{c|}{Color Robustness} & \multicolumn{2}{c|}{Text Robustness} & \multicolumn{2}{c|}{Layout Robustness} & \multirow{2}{*}{Safety Critical} \\
    \cline{2-4} \cline{6-7} \cline{9-12} \cline{13-14} \cline{15-16}
    & easy & medium & hard &  & easy & hard &  & before & darkening & 10--30\% & 100\% & before & after & before & after  &  \\
    \hline
    \rowcolor{grayheader}
    \multicolumn{17}{l}{\it\textbf{closed Source Model}} \\
    GPT-5 & \textbf{63.5} & \textbf{50.0} & 41.3 & 91.9 & \textbf{51.0} & 32.6 & \textbf{83.4} & 72.6 & 47.3(\textcolor{red}{$\downarrow$25.3}) & \textbf{66.5}(\textcolor{red}{$\downarrow$6.1}) & 54.9(\textcolor{red}{$\downarrow$17.7}) & 57.9 & 53.6(\textcolor{red}{$\downarrow$4.3}) & 18.5 & 18.2(\textcolor{red}{$\downarrow$0.3}) & 71.1 \\
    claude-4-Sonnet & 23.7 & 16.6 & 14.5 & 88.7 & 30.3 & 31.8 & 72.1 & 65.2 & 36.2(\textcolor{red}{$\downarrow$29.0}) & 56.5(\textcolor{red}{$\downarrow$8.7}) & 46.6(\textcolor{red}{$\downarrow$18.6}) & 52.5 & 46.6(\textcolor{red}{$\downarrow$5.9}) & 34.6 & 32.0(\textcolor{red}{$\downarrow$2.6}) & 53.3 \\
    Gemini 2.5-Pro & 56.1 & 49.5 & \textbf{44.2} & \textbf{92.1} & 41.5 & 29.1 & 77.2 & 71.4 & 45.4(\textcolor{red}{$\downarrow$26.0}) & 62.7(\textcolor{red}{$\downarrow$8.7}) & 53.9(\textcolor{red}{$\downarrow$17.5}) & 58.0 & 53.5(\textcolor{red}{$\downarrow$4.5}) & 28.7 & 27.5(\textcolor{red}{$\downarrow$1.2}) & \textbf{91.1} \\
    \hline
    \rowcolor{grayheader}
    \multicolumn{17}{l}{\it\textbf{Open Source Model}} \\
    Pixtral-Large & 17.4 & 13.2 & 10.9 & 85.3 & 33.7 & 21.4 & 65.0 & 38.7 & 19.8(\textcolor{red}{$\downarrow$18.9}) & 31.9(\textcolor{red}{$\downarrow$6.8}) & 27.2(\textcolor{red}{$\downarrow$11.5}) & 48.1 & 42.2(\textcolor{red}{$\downarrow$5.9}) & 43.5 & 38.4(\textcolor{red}{$\downarrow$5.2}) & 75.6 \\
    Qwen 2.5-VL-72B & 30.7 & 25.4 & 20.0 & 85.3 & 38.8 & 31.9 & 73.2 & \textbf{72.8} & \textbf{47.6}(\textcolor{red}{$\downarrow$25.2}) & 64.4(\textcolor{red}{$\downarrow$8.4}) & \textbf{55.9}(\textcolor{red}{$\downarrow$16.9}) & \textbf{80.1} & \textbf{70.8(\textcolor{red}{$\downarrow$9.3})} & 47.4 & \textbf{42.5}(\textcolor{red}{$\downarrow$4.9}) & 71.1 \\
    \hline
    Qwen-vl-plus & 18.2 & 21.1 & 19.8 & 72.1 & 40.2 & 32.7 & 68.3 & 61.9 & 41.1(\textcolor{red}{$\downarrow$20.8}) & 57.8(\textcolor{red}{$\downarrow$4.1}) & 51.7(\textcolor{red}{$\downarrow$10.2}) & 65.1 & 56.1(\textcolor{red}{$\downarrow$9.0}) & 49.6 & 42.1(\textcolor{red}{$\downarrow$7.5}) & 75.6 \\
    Qwen2.5-VL-32B & 13.7 & 12.5 & 12.1 & 87.5 & 40.3 & 26.9 & 60.5 & 40.8 & 43.3(\textcolor{green}{$\uparrow$2.5}) & 37.6(\textcolor{red}{$\downarrow$3.2}) & 49.9(\textcolor{green}{$\uparrow$9.1}) & 67.6 & 59.7(\textcolor{red}{$\downarrow$7.9}) & 37.3 & 30.5(\textcolor{red}{$\downarrow$6.8}) & 80.0 \\
    Intern-S1 & 21.7 & 9.3 & 11.3 & 81.5 & 42.4 & 33.0 & 73.5 & 69.8 & 41.9(\textcolor{red}{$\downarrow$27.9}) & 62.8(\textcolor{red}{$\downarrow$7.0}) & 49.0(\textcolor{red}{$\downarrow$20.8}) & 73.0 & 64.0(\textcolor{red}{$\downarrow$9.0}) & 47.3 & 40.9(\textcolor{red}{$\downarrow$6.4}) & 80.0 \\
    \hline
    Llama4-Scout17B & 23.0 & 23.3 & 26.7 & 82.1 & 42.3 & 29.3 & 64.5 & 62.8 & 39.4(\textcolor{red}{$\downarrow$23.4}) & 56.8(\textcolor{red}{$\downarrow$6.0}) & 48.5(\textcolor{red}{$\downarrow$14.3}) & 36.6 & 29.9(\textcolor{red}{$\downarrow$6.7}) & 38.8 & 35.7(\textcolor{red}{$\downarrow$3.1}) & 86.7 \\
    Pixtral-12B & 15.5 & 15.5 & 16.1 & 60.6 & 35.8 & 32.0 & 52.3 & 35.8 & 18.4(\textcolor{red}{$\downarrow$17.4}) & 30.3(\textcolor{red}{$\downarrow$5.5}) & 25.7(\textcolor{red}{$\downarrow$10.1}) & 29.4 & 25.2(\textcolor{red}{$\downarrow$4.2}) & 32.4 & 31.6(\textcolor{red}{$\downarrow$0.8}) & 40.0 \\
    Qwen2.5-VL-7B & 4.7 & 5.0 & 6.8 & 67.6 & 41.0 & 31.7 & 64.5 & 60.8 & 42.3(\textcolor{red}{$\downarrow$18.5}) & 56.7(\textcolor{red}{$\downarrow$4.1}) & 49.8(\textcolor{red}{$\downarrow$11.0}) & 50.9 & 42.4(\textcolor{red}{$\downarrow$8.5}) & 24.7 & 23.7(\textcolor{red}{$\downarrow$1.1}) & \textbf{91.1} \\
    \bottomrule
  \end{tabular}}
\end{table*}

\begin{figure*}[ht]
  \includegraphics[width=\linewidth]{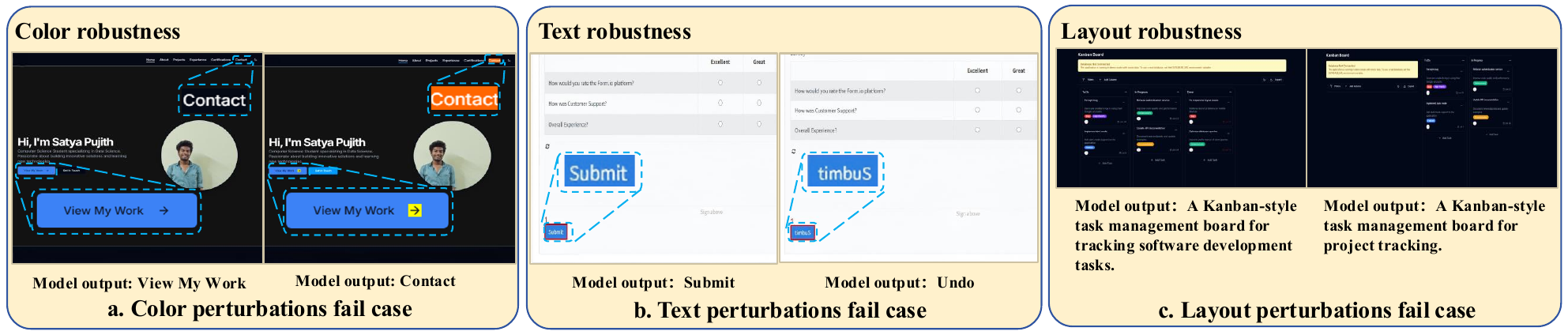}
  \caption{Output examples of MLLMs on \bench \ when facing different perturbations.}
  \label{fig:case}
\end{figure*}

Overall, the table underscores the heterogeneity of model capabilities: while progress in robustness is notable, reasoning and safety-critical understanding continue to pose challenges, preventing any single model from dominating across all eight dimensions.\\


\begin{tcolorbox}[colback=gray!20, colframe=gray!20, width=\columnwidth, left=0.05in, right=0.05in, top=0.05in, bottom=0.05in]
\textbf{Answer to RQ1:} Closed-source models consistently outperform open-source alternatives across all dimensions, particularly in safety tasks. Larger open-source models show competitive robustness performance but struggle with reasoning tasks like position relationship and form filling, which remain the most challenging for all MLLMs.
\end{tcolorbox}

\subsection{Fine-tuning results (RQ2)}
We next examine the impact of fine-tuning on model performance. For the pairwise relative position reasoning task, the Qwen2.5-VL-7B originally achieved only around 16.3\%\ accuracy. After fine-tuning, the accuracy increased substantially to 41.3\%\, showing a clear improvement in spatial reasoning ability. On the UI-group task, the model’s accuracy rose from 67.6\%\ before fine-tuning to 96.9\%\ afterwards, indicating that targeted supervision allowed the model to capture structural layout patterns more effectively. For color robustness, the baseline average accuracy across easy, medium, and hard settings was 73.1\%\, while fine-tuning raised the average accuracy to 80.1\%\. These results collectively demonstrate that targeted LoRA-based fine-tuning yields consistent and notable gains across multiple benchmark dimensions.\\

\begin{tcolorbox}[colback=gray!20, colframe=gray!20, width=\columnwidth, left=0.05in, right=0.05in, top=0.05in, bottom=0.05in]
\textbf{Answer to RQ2:} LoRA-based fine-tuning significantly improves model performance across all tasks, with notable gains in position reasoning (16.3\% → 41.3\%), UI-grouping (67.6\% → 96.9\%), and color robustness (73.1\% → 80.1\%), demonstrating that targeted supervision effectively enhances MLLMs' web understanding capabilities.
\end{tcolorbox}

\subsection{Failure Reason Analysis (RQ3)}

We summarize representative failure cases to answer RQ3. Although our perturbations are designed to preserve the underlying semantics of webpages, models often produce inconsistent outputs as shown in Figure~\ref{fig:case}. Such failures arise from systematic weaknesses that manifest differently under color, text, and layout perturbations. (1) Color perturbations lead to two major types of failures: highly saturated or visually striking colors dominate attention, causing predictions to rely on color blocks rather than textual or structural cues, and background recoloring reduces text–background contrast, degrading OCR and producing incorrect button text. (2) The limited ability between character-level OCR recognition and holistic page reasoning makes models brittle: even minor character edits  cause large deviations in functional interpretation. (3) Layout perturbations often drives models to over-attend to a single region while neglecting the global structure, resulting in incomplete or oversimplified summaries of webpage functionality. The detailed Models' outputs self comparison and fail cases in our repo~\cite{WebRSSBench}.


\begin{tcolorbox}[colback=gray!20, colframe=gray!20, width=\columnwidth, left=0.05in, right=0.05in, top=0.05in, bottom=0.05in]
\textbf{Answer to RQ3:} Models fail due to three systematic weaknesses: color perturbations cause over-reliance on visual salience and degrade OCR through reduced contrast; text perturbations reveal brittleness in character-level recognition where minor edits cause large deviations in functional interpretation; layout perturbations drive models to over-attend to single regions while neglecting global structure, resulting in incomplete functionality summaries.
\end{tcolorbox}


\section{Conclusion}
In this paper, we introduce \bench, a large-scale and comprehensive benchmark for evaluating the reasoning ability, robustness, and safety of multimodal large language models (MLLMs) in web-related tasks. \bench\ consists of 729 webpages and 3799 QA pairs, covering eight tasks, and involves the evaluation of twelve open-source and closed-source models. Our results reveal notable shortcomings of current models in handling web tasks, as well as the lack of systematic research on robustness and safety. This benchmark is intended to provide representative examples for future studies on reasoning, robustness, and safety, where the designed perturbation strategies align with contemporary jailbreak research and offer concrete test cases. Ultimately, \bench\ provides insights and guidance for advancing the next generation of models toward stronger reasoning, greater robustness, and improved safety.




\bibliography{ref}

@misc{xiao2026comuicoder,
      title={ComUICoder: Component-based Reusable UI Code Generation for Complex Websites via Semantic Segmentation and Element-wise Feedback}, 
      author={Jingyu Xiao and Jiantong Qin and Shuoqi Li and Man Ho Lam and Yuxuan Wan and Jen-tse Huang and Yintong Huo and Michael R. Lyu},
      year={2026},
      eprint={2602.19276},
      archivePrefix={arXiv},
      primaryClass={cs.SE},
      url={https://arxiv.org/abs/2602.19276}, 
}

@article{wan2025automatically,
  title={Automatically Generating Web Applications from Requirements Via Multi-Agent Test-Driven Development},
  author={Wan, Yuxuan and Liang, Tingshuo and Xu, Jiakai and Xiao, Jingyu and Huo, Yintong and Lyu, Michael R},
  journal={arXiv preprint arXiv:2509.25297},
  year={2025}
}

@article{chen2025survey,
  title={A Survey on the Safety and Security Threats of Computer-Using Agents: JARVIS or Ultron?},
  author={Chen, Ada and Wu, Yongjiang and Zhang, Junyuan and Xiao, Jingyu and Yang, Shu and Huang, Jen-tse and Wang, Kun and Wang, Wenxuan and Wang, Shuai},
  journal={arXiv preprint arXiv:2505.10924},
  year={2025}
}

@misc{v0community,
  author       = {Vercel},
  title        = {V0 Community},
  year         = {2025},
  howpublished = {\url{https://v0.app/templates}},
  note         = {Accessed: 2025-09-25}
}

@inproceedings{yue2024mmmu,
  title={Mmmu: A massive multi-discipline multimodal understanding and reasoning benchmark for expert agi},
  author={Yue, Xiang and Ni, Yuansheng and Zhang, Kai and Zheng, Tianyu and Liu, Ruoqi and Zhang, Ge and Stevens, Samuel and Jiang, Dongfu and Ren, Weiming and Sun, Yuxuan and others},
  booktitle={Proceedings of the IEEE/CVF Conference on Computer Vision and Pattern Recognition},
  pages={9556--9567},
  year={2024}
}

@article{awal2024webmmu,
  title = {WebMMU: A Benchmark for Multimodal Multilingual Website Understanding and Code Generation},
  author = {Rabiul Awal and Mahsa Massoud and Zichao Li and Aarash Feizi and Suyuchen Wang and Christopher Pal and Aishwarya Agrawal and David Vazquez and Siva Reddy and Juan A. Rodriguez and Perouz Taslakian and Spandana Gella and Sai Rajeswar},
  journal = {arXiv preprint arXiv:2406.05018},
  year = {2024},
  url = {https://arxiv.org/abs/2406.05018}
}

@article{wang2024design2code,
  title={Design2Code: Benchmarking Multimodal Code Generation for Automated Front-End Engineering},
  author={Wang, Sihua and Gao, Yifan and Sun, Qian and Fan, Xin and Liu, Xianjun and Yang, Song and Tang, Jian},
  journal={arXiv preprint arXiv:2406.03096},
  year={2024},
  url={https://arxiv.org/abs/2406.03096}
}

@inproceedings{deng2024mind2web,
  title={Mind2Web: Towards a Generalist Agent for the Web},
  author={Deng, Xiang and Gu, Yu and Zheng, Boyuan and Chen, Shijie and Stevens, Sam and Wang, Boshi and Sun, Huan and Su, Yu},
  booktitle={International Conference on Learning Representations (ICLR)},
  year={2024},
  url={https://openreview.net/forum?id=40QXyfcVn4}
}

@inproceedings{lin-etal-2025-webuibench,
  title        = "{W}eb{UIB}ench: A Comprehensive Benchmark for Evaluating Multimodal Large Language Models in {W}eb{UI}-to-Code",
  author       = "Lin, Zhiyu and Zhou, Zhengda and Zhao, Zhiyuan and Wan, Tianrui and Ma, Yilun and Gao, Junyu and Li, Xuelong",
  booktitle    = "Findings of the Association for Computational Linguistics: ACL 2025",
  editor       = "Che, Wanxiang and Nabende, Joyce and Shutova, Ekaterina and Pilehvar, Mohammad Taher",
  month        = jul,
  year         = "2025",
  address      = "Vienna, Austria",
  publisher    = "Association for Computational Linguistics",
  pages        = "15780--15797",
  doi          = "10.18653/v1/2025.findings-acl.815",
}

@article{liu2023mmbench,
  title   = {MMBench: Is Your Multi-Modal Model an All-Around Player?},
  author  = {Liu, Yuan and Duan, Haodong and Zhang, Yuanhan and Li, Bo and Zhang, Songyang and Zhao, Wangbo and Yuan, Yike and Wang, Jiaqi and He, Conghui and Liu, Ziwei and others},
  journal = {arXiv preprint arXiv:2307.06281},
  year    = {2023}
}

@misc{liu2024visualwebbench,
  title        = {VisualWebBench: How Far Have Multimodal LLMs Evolved in Web Page Understanding and Grounding?},
  author       = {Liu, Junpeng and Song, Yifan and Lin, Bill Yuchen and Lam, Wai and Neubig, Graham and Li, Yuanzhi and Yue, Xiang},
  year         = {2024},
  eprint       = {2404.05955},
  archivePrefix = {arXiv},
  primaryClass = {cs.CL}
}

@article{xiao2025efficientuicoder,
  title={EfficientUICoder: Efficient MLLM-based UI Code Generation via Input and Output Token Compression},
  author={Xiao, Jingyu and Zhang, Zhongyi and Wan, Yuxuan and Huo, Yintong and Liu, Yang and Lyu, Michael R},
  journal={arXiv preprint arXiv:2509.12159},
  year={2025}
}

@inproceedings{wu-etal-2025-webwalker,
  title     = "{W}ebWalker: Benchmarking LLMs in Web Traversal",
  author    = "Wu, Jialong and Yin, Wenbiao and Jiang, Yong and Wang, Zhenglin and Xi, Zekun and Fang, Runnan and Zhang, Linhai and He, Yulan and Zhou, Deyu and Xie, Pengjun and Huang, Fei",
  booktitle = "Proceedings of the 63rd Annual Meeting of the Association for Computational Linguistics (ACL 2025)",
  year      = "2025",
  address   = "Vienna, Austria",
  publisher = "Association for Computational Linguistics"
}

@inproceedings{xu-etal-2025-turkingbench,
  title     = "{T}urking{B}ench: A Challenge Benchmark for Web Agents",
  author    = "Xu, Kevin and Kordi, Yeganeh and Nayak, Tanay and Asija, Adi and Wang, Yizhong and Sanders, Kate and Byerly, Adam and Zhang, Jingyu and Van Durme, Benjamin and Khashabi, Daniel",
  editor    = "Chiruzzo, Luis and Ritter, Alan and Wang, Lu",
  booktitle = "Proceedings of the 2025 Conference of the Nations of the Americas Chapter of the Association for Computational Linguistics: Human Language Technologies (Volume 1: Long Papers)",
  month     = apr,
  year      = "2025",
  address   = "Albuquerque, New Mexico",
  publisher = "Association for Computational Linguistics",
  pages     = "3694--3710",
  doi       = "10.18653/v1/2025.naacl-long.188"
}

@article{tang2025slidecoder,
  title={SlideCoder: Layout-aware RAG-enhanced Hierarchical Slide Generation from Design},
  author={Tang, Wenxin and Xiao, Jingyu and Jiang, Wenxuan and Xiao, Xi and Wang, Yuhang and Tang, Xuxin and Li, Qing and Ma, Yuehe and Liu, Junliang and Tang, Shisong and others},
  journal={arXiv preprint arXiv:2506.07964},
  year={2025}
}

@inproceedings{gui2025webcode2m,
  title={Webcode2m: A real-world dataset for code generation from webpage designs},
  author={Gui, Yi and Li, Zhen and Wan, Yao and Shi, Yemin and Zhang, Hongyu and Chen, Bohua and Su, Yi and Chen, Dongping and Wu, Siyuan and Zhou, Xing and others},
  booktitle={Proceedings of the ACM on Web Conference (WWW)},
  pages={1834--1845},
  year={2025}
}

@article{xiao2025designbench,
  title={Designbench: A comprehensive benchmark for mllm-based front-end code generation},
  author={Xiao, Jingyu and Wang, Ming and Lam, Man Ho and Wan, Yuxuan and Liu, Junliang and Huo, Yintong and Lyu, Michael R},
  journal={arXiv preprint arXiv:2506.06251},
  year={2025}
}

@inproceedings{si2025design2code,
    title = "Design2Code: Benchmarking Multimodal Code Generation for Automated Front-End Engineering",
    author = "Si, Chenglei and Zhang, Yanzhe and Li, Ryan  and Yang, Zhengyuan  and Liu, Ruibo and Yang, Diyi",
    booktitle = "Proceedings of the 2025 Conference of the Nations of the Americas Chapter of the Association for Computational Linguistics: Human Language Technologies (NAACL)",
    month = apr,
    year = "2025",
    address = "Albuquerque, New Mexico",
    publisher = "Association for Computational Linguistics",
    pages = "3956--3974",
    ISBN = "979-8-89176-189-6"
}

@inproceedings{yun2024web2code,
 author = {Yun, Sukmin and Lin, Haokun and Thushara, Rusiru and Bhat, Mohammad Qazim and Wang, Yongxin and Jiang, Zutao and Deng, Mingkai and Wang, Jinhong and Tao, Tianhua and Li, Junbo and Li, Haonan and Nakov, Preslav and Baldwin, Timothy and Liu, Zhengzhong and Xing, Eric P. and Liang, Xiaodan and Shen, Zhiqiang},
 booktitle = {Advances in Neural Information Processing Systems (NeurIPS)},
 pages = {112134--112157},
 publisher = {Curran Associates, Inc.},
 title = {Web2Code: A Large-scale Webpage-to-Code Dataset and Evaluation Framework for Multimodal LLMs},
 volume = {37},
 year = {2024}
}

@article{wan2024mrweb,
  title={MRWeb: An Exploration of Generating Multi-Page Resource-Aware Web Code from UI Designs},
  author={Wan, Yuxuan and Dong, Yi and Xiao, Jingyu and Huo, Yintong and Wang, Wenxuan and Lyu, Michael R},
  journal={arXiv preprint arXiv:2412.15310},
  year={2024}
}

@article{xiao2024interaction2code1,
  title={Interaction2Code: Benchmarking MLLM-based Interactive Webpage Code Generation from Interactive Prototyping},
  author={Xiao, Jingyu and Wan, Yuxuan and Huo, Yintong and Wang, Zixin and Xu, Xinyi and Wang, Wenxuan and Xu, Zhiyao and Wang, Yuhang and Lyu, Michael R},
  journal={arXiv preprint arXiv:2411.03292},
  year={2024}
}

@misc{laurençon2024unlocking,
      title={Unlocking the conversion of Web Screenshots into HTML Code with the WebSight Dataset}, 
      author={Hugo Laurençon and Léo Tronchon and Victor Sanh},
      year={2024},
      eprint={2403.09029},
      archivePrefix={arXiv},
      primaryClass={cs.HC}
}

@misc{wcag21,
  title        = {Web Content Accessibility Guidelines (WCAG) 2.1},
  author       = {{World Wide Web Consortium (W3C)}},
  year         = {2018},
  howpublished = {\url{https://www.w3.org/TR/WCAG21/}},
  note         = {W3C Recommendation}
}

@misc{openai_gpt5_2025,
  title        = {GPT-5},
  author       = {OpenAI},
  howpublished = {\url{https://openai.com}},
  year         = {2025},
  note         = {Official model page / blog; use the specific release URL if available}
}

@misc{anthropic_claude4_sonnet_2025,
  title        = {Claude 4 Sonnet},
  author       = {Anthropic},
  howpublished = {\url{https://www.anthropic.com}},
  year         = {2025},
  note         = {Model card / product page}
}

@misc{deepmind_gemini25_pro_2025,
  title        = {Gemini 2.5 Pro},
  author       = {Google DeepMind},
  howpublished = {\url{https://deepmind.google/technologies/gemini/}},
  year         = {2025},
  note         = {Product page; for technical details also cite the Gemini 1.5 technical report}
}

@misc{mistral_pixtral_large_2025,
  title        = {Pixtral-Large},
  author       = {Mistral AI},
  howpublished = {\url{https://mistral.ai}},
  year         = {2025},
  note         = {Release / blog announcement}
}

@misc{qwen25_vl_72b_2025,
  title        = {Qwen2.5-VL-72B},
  author       = {Alibaba Cloud and Qwen Team},
  howpublished = {\url{https://github.com/QwenLM/Qwen2.5-VL}},
  year         = {2025},
  note         = {Repository / docs}
}

@misc{qwen_vl_plus_2025,
  title        = {Qwen-VL-Plus},
  author       = {Alibaba Cloud and Qwen Team},
  howpublished = {\url{https://qwenlm.github.io}},
  year         = {2025},
  note         = {API variant / model card}
}

@misc{qwen25_vl_32b_2025,
  title        = {Qwen2.5-VL-32B},
  author       = {Alibaba Cloud and Qwen Team},
  howpublished = {\url{https://github.com/QwenLM/Qwen2.5-VL}},
  year         = {2025},
  note         = {Repository / docs}
}

@misc{intern_s1_2025,
  title        = {Intern-S1},
  author       = {Shanghai AI Lab},
  howpublished = {\url{https://github.com/InternLM}},
  year         = {2025},
  note         = {Project / model card}
}

@misc{groq_llama4_scout17b_2025,
  title        = {Llama4-Scout-17B},
  author       = {Groq},
  howpublished = {\url{https://groq.com}},
  year         = {2025},
  note         = {Release notes / model page}
}

@misc{mistral_pixtral_12b_2024,
  title        = {Pixtral-12B},
  author       = {Mistral AI},
  howpublished = {\url{https://mistral.ai}},
  year         = {2024},
  note         = {Release / blog announcement}
}

@misc{qwen25_vl_7b_2025,
  title        = {Qwen2.5-VL-7B},
  author       = {Alibaba Cloud and Qwen Team},
  howpublished = {\url{https://github.com/QwenLM/Qwen2.5-VL}},
  year         = {2025},
  note         = {Repository / docs}
}

@misc{moz_top500,
  title        = {The Moz Top 500 Most Popular Websites},
  howpublished = {\url{https://moz.com/top500}},
  note         = {Accessed: 2025-07-01},
  author       = {{Moz}},
  year         = {2025}
}

@article{zou2025queryattack,
  title={QueryAttack: Jailbreaking Aligned Large Language Models Using Structured Non-natural Query Language},
  author={Zou, Qingsong and Xiao, Jingyu and Li, Qing and Yan, Zhi and Wang, Yuhang and Xu, Li and Wang, Wenxuan and Gao, Kuofeng and Li, Ruoyu and Jiang, Yong},
  journal={arXiv preprint arXiv:2502.09723},
  year={2025}
}

@inproceedings{chen-etal-2021-websrc,
  title     = {WebSRC: A Dataset for Web-Based Structural Reading Comprehension},
  author    = {Chen, Xingyu and Zhao, Zihan and Chen, Lu and Jiabao Ji and Zhang, Danyang and Luo, Ao and Xiong, Yuxuan and Yu, Kai},
  booktitle = {Proceedings of the 2021 Conference on Empirical Methods in Natural Language Processing},
  month     = nov,
  year      = {2021},
  address   = {Online and Punta Cana, Dominican Republic},
  publisher = {Association for Computational Linguistics},
  url       = {https://aclanthology.org/2021.emnlp-main.343/},
  doi       = {10.18653/v1/2021.emnlp-main.343},
  pages     = {4173--4185}
}

@inproceedings{chen2021webqa,
  title={WebQA: Multihop and Multimodal QA},
  author={Chen, Wenhu and Zha, Hanwen and Chen, Zhiyu and Yan, Xiyou and Su, Yu and Wang, William and Yih, Wen-tau and Zhang, Yizhou},
  booktitle={Thirty-fifth Conference on Neural Information Processing Systems (NeurIPS) Datasets and Benchmarks Track},
  year={2021}
}

@inproceedings{awal2024vismin,
  title={VisMin: Visual Minimal-Change Understanding},
  author={Awal, Rabiul and Ahmadi, Saba and Zhang, Le and Agrawal, Aishwarya},
  booktitle={Advances in Neural Information Processing Systems},
  volume={37},
  pages={10234--10258}, 
  year={2024}
}

@article{feizi2025pairbench,
  title={PairBench: A Systematic Framework for Selecting Reliable Judge VLMs},
  author={Feizi, Aarash and Rajeswar, Sai and Romero-Soriano, Adriana and Rabbany, Reihaneh and Gella, Spandana and Zantedeschi, Valentina and Monteiro, Jo{\~a}o},
  journal={arXiv preprint arXiv:2502.15210},
  year={2025}
}

@inproceedings{nguyen2025gui,
  title={Gui agents: A survey},
  author={Nguyen, Dang and Chen, Jian and Wang, Yu and Wu, Gang and Park, Namyong and Hu, Zhengmian and Lyu, Hanjia and Wu, Junda and Aponte, Ryan and Xia, Yu and others},
  booktitle={Findings of the Association for Computational Linguistics: ACL 2025},
  pages={22522--22538},
  year={2025}
}

@inproceedings{hong2024cogagent,
  title={Cogagent: A visual language model for gui agents},
  author={Hong, Wenyi and Wang, Weihan and Lv, Qingsong and Xu, Jiazheng and Yu, Wenmeng and Ji, Junhui and Wang, Yan and Wang, Zihan and Dong, Yuxiao and Ding, Ming and others},
  booktitle={Proceedings of the IEEE/CVF Conference on Computer Vision and Pattern Recognition},
  pages={14281--14290},
  year={2024}
}

@article{zhang2024large,
  title={Large language model-brained gui agents: A survey},
  author={Zhang, Chaoyun and He, Shilin and Qian, Jiaxu and Li, Bowen and Li, Liqun and Qin, Si and Kang, Yu and Ma, Minghua and Liu, Guyue and Lin, Qingwei and others},
  journal={arXiv preprint arXiv:2411.18279},
  year={2024}
}

@misc{WebRSSBench,
  title        = {Detail Infomation of out {WebRSSBench}},
  author       = {{Annoy-worker}},
  year         = {2025},
  howpublished = {\url{https://github.com/annoy-worker/WebRSSBench}},
  note         = {Accessed: 2026-02-11}
}
\bibliographystyle{ACM-Reference-Format}

\end{document}